# Nuevo Indice de Similitud para Imágenes basado en Entropía y Teoría de Grupos

Y. Garcés, E. Torres, O. Pereira and R. Rodríguez

*Abstract*— **In this work, we propose a new similarity index for images considering the entropy function and group theory. This index considers an algebraic group of images, it is defined by an inner law that provides a novel approach for the subtraction of images. Through an equivalence relationship in the field of images, we prove the existence of the quotient group, on which the new similarity index is defined. We also present the main properties of the new index, and the immediate application thereof as a stopping criterion of the ``Mean Shift Iterative Algorithm''.**

## INTRODUCCION

La segmentación es la partición de una imagen en un subconjunto de regiones homogéneas en base a una característica dada, por ejemplo, intensidad o textura. El objetivo de la segmentación es simplificar y/o cambiar la representación de una imagen en otra más significativa y fácil de analizar.

Algunos de los principales algoritmos de segmentación utilizan una delimitación minuciosa de los objetos presentes en la imagen a través de un observador experimentado. Este tipo de procedimiento se denomina comúnmente como segmentación manual. Dicha práctica constituye una tarea tediosa y poco repetible, además de ser propensa a errores. Por este motivo el desarrollo de métodos de segmentación que sean independientes de la intervención del usuario contribuye sustancialmente a la precisión de la segmentación de imágenes. Estos métodos se denominan automáticos y el objetivo fundamental es facilitar el uso rutinario por parte del usuario entrenado y no entrenado, evitando disparidad de criterios.

En nuestros días, no existe una teoría general para la segmentación de imágenes, aunque se ha dedicado mucho esfuerzo en desarrollar estrategias de segmentación. Tradicionalmente el desarrollo de algoritmos de segmentación ha sido un proceso orientado a las aplicaciones, de forma que los mismos funcionan correctamente solo en algunos tipos de imágenes. Como consecuencia, se han explorado muchas direcciones de investigación y han sido adoptados principios muy diferentes, por lo que han aparecido en la literatura una amplia variedad de algoritmos de segmentación con variadas bases teóricas [6, 11, 15].

Un algoritmo que ha alcanzado relativa importancia en la actualidad para la segmentación de imágenes es el "Algoritmo Iterativo de la Media Desplazada" ($MSHi$) [6, 8, 9, 10, 11]. El mismo considera el espacio de características de una imagen como una muestra de una función de densidad de probabilidad desconocida, que se estima mediante un método no paramétrico basado en el kernel, el cual opera en el dominio rango-espacial. Como este algoritmo es no supervisado se impone la necesidad de establecer un criterio de parada consecuente con el nivel de segmentación que el usuario desea. Uno de los enfoques empleados para obtener dicho criterio de parada se basa en considerar la distancia entre las imágenes consecutivas que se obtienen como parte del proceso iterativo de segmentación [8, 9, 10]. Sin embargo, establecer una distancia adecuada entre imágenes de forma que se obtenga eficiencia computacional y confiabilidad en el nivel de discrepancia entre las mismas no es una tarea sencilla y constituye una de las áreas de investigación más activa en la actualidad [1, 2, 3, 4].

En el paradigma convencional [5], la medición de la similitud entre imágenes implica la selección de aquellas propiedades consideradas como importantes. Estas subjetivas decisiones son erróneas en ocasiones y siempre implican cierta pérdida de información sobre los datos originales. Por esta razón, en ocasiones es preferible utilizar técnicas que permitan emplear toda la información disponible sobre las entidades involucradas.

Las distintas medidas de similitud introducidas hasta la fecha pueden ser sensibles a variadas situaciones como modalidad, contenido y diferencia de la imagen en cuanto a muestreo, interpolación, solapamiento parcial y degradación (por ejemplo ruido) [20]. Para elegir la medida de similitud y su implementación de manera apropiada, es a menudo deseable disponer de alguna información a priori sobre el comportamiento de la función de similitud con respecto a los factores mencionados y al conjunto de imágenes donde se desea emplear dicho índice.

En el presente trabajo, se propone un nuevo índice de similitud entre imágenes basado en la Teoría de Grupos y la función de entropía de Shannon. Por las características del nuevo índice de similitud el mismo se aplica inicialmente como un nuevo criterio de parada para el "Algoritmo Iterativo de la Media Desplazada", y se demuestra teórica y experimentalmente que con este nuevo criterio se obtiene mayor estabilidad en el proceso de segmentación que con el criterio usado por los autores en [8, 9, 10]. Además, se realiza un estudio sobre las potencialidades del nuevo índice de similitud como un nuevo enfoque para evaluar la segmentación de imágenes digitales.

Y. Garcés, Instituto de Cibernética Matemática y Física, La Habana, Cuba, 88yasel@gmail.com

E. Torres, Instituto de Cibernética Matemática y Física, La Habana, Cuba, esley@icimaf.cu

O. Pereira, Instituto de Cibernética Matemática y Física, La Habana, Cuba, opererira@icimaf.cu

R. Rodríguez, Instituto de Cibernética Matemática y Física, La Habana, Cuba, rrm@icimaf.cu

## I. CONSIDERACIONES SOBRE EL ALGORITMO $MSHi$

En esta sección se estudia el índice de similitud empleado en [7, 8, 9, 10] como criterio de parada del algoritmo $MSHi$.

En [7, 8, 9, 10] la distancia entre dos imágenes A y B se toma de acuerdo a la siguiente ecuación:
$$v(A,B) = |E(A) - E(B)|, \qquad (1)$$

donde $E(\cdot)$ es la función de entropía de Shannon [15]. El algoritmo $MSHi$ se detiene cuando $v(A_k, A_{k-1}) \leq \varepsilon$, $A_k$ denota la imagen obtenida por el proceso de segmentación en el paso $k$, mientras que $\varepsilon$ es el umbral de parada establecido.

La Figura 1 muestra la segmentación de la imagen clásica de "Bárbara" al aplicar el algoritmo $MSHi$ con el criterio de parada (1). En Figura 1(c) se ha graficado el comportamiento del criterio de parada (eje de las ordenadas), en contraste con el número de iteraciones que realiza el algoritmo (eje de las abscisas). Se puede apreciar como la curva que describe el criterio de parada no es suave; o sea, el índice en la expresión (1) no capta la diferencia real que existe entre las imágenes obtenidas durante el proceso de segmentación. Esto puede provocar que el algoritmo no se detenga a tiempo ocasionando sobresegmentación en la imagen, o al contrario, pare muy rápido, por lo que la imagen no quedaría debidamente segmentada, según el criterio del observador.

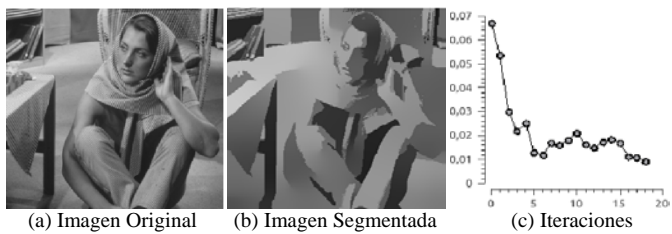

(a) Imagen Original  (b) Imagen Segmentada  (c) Iteraciones

Figura 1. Ejemplo de segmentación usando $MSHi$.

**Definición 1.1 (Equivalencia Débil)** Dos imágenes $A$ y $B$ se dicen débilmente equivalentes si $E(A) = E(B)$. Esta equivalencia se denota como $A \asymp B$.

Es sencillo demostrar que la Definición 1.1 establece una relación de equivalencia en el campo de las imágenes. Un resultado inmediato es que dos imágenes $A$ y $B$ son débilmente equivalentes si y solo si $v(A,B) = 0$, o sea, el criterio dado en la expresión (1) es una caracterización de la equivalencia débil. El problema de este criterio es que no considera la información espacial entre las imágenes sino solamente la frecuencia de los niveles de grises, por lo que, dos imágenes diferentes pueden tener mediante el índice de similitud en la expresión (1) un valor muy pequeño. Esta es una razón de peso para considerar que el criterio de parada definido en la expresión (1) no es apropiado y proporciona inestabilidad en el algoritmo $MSHi$.

Justamente, al ser el algoritmo $MSHi$ no supervisado, es imprescindible que posea un criterio de parada que logre captar la diferencia real que existe entre las imágenes, proporcionando de esta forma mayor estabilidad en el proceso de segmentación.

## II. SUSTRACCIÓN DE IMÁGENES

Es lógico considerar que existe poca diferencia entre dos imágenes si al sustraerlas la imagen resultante está próxima a la imagen nula (imagen con intensidad cero en todos los píxeles). Esta idea a pesar de su sencillez, presenta problemas debido a que las imágenes están definidas en $\mathbb{Z}_+$, o sea, en los enteros positivos, y mediante la sustracción se pueden generar valores negativos en algunos píxeles. En el Ejemplo 1 se muestra el resultado obtenido al restar dos imágenes.

**Ejemplo 1** Sustracción de imágenes mediante la resta usual en $\mathbb{R}$. Como se observa el resultado de dicha operación no es una imagen pues algunos píxeles tienen valores negativos en las intensidades de los niveles de gris.

$$\begin{pmatrix} 8 & 3 & 2 \\ 9 & 15 & 1 \\ 4 & 7 & 2 \end{pmatrix} - \begin{pmatrix} 8 & 1 & 5 \\ 3 & 12 & 2 \\ 6 & 4 & 1 \end{pmatrix} = \begin{pmatrix} 0 & 2 & -3 \\ 6 & 3 & -1 \\ -2 & 3 & 1 \end{pmatrix}$$

Las alternativas usadas hasta el momento en el campo del procesamiento de imágenes para resolver este problema son:

1. *Truncamiento*: Los píxeles que quedan fuera del rango de definición de la imagen diferencia se truncan en cero. En este caso el problema del Ejemplo 1 queda resuelto de la siguiente forma:
$$\begin{pmatrix} 8 & 3 & 2 \\ 9 & 15 & 1 \\ 4 & 7 & 2 \end{pmatrix} - \begin{pmatrix} 8 & 1 & 5 \\ 3 & 12 & 2 \\ 6 & 4 & 1 \end{pmatrix} = \begin{pmatrix} 0 & 2 & 0 \\ 6 & 3 & 0 \\ 0 & 3 & 1 \end{pmatrix}$$

   La deficiencia del truncamiento es que se pierde información importante en la imagen diferencia y se generan nuevas frecuencias del valor cero cuando realmente no debiesen existir. Al hacer cero los valores negativos implícitamente se asume que las imágenes $A$ y $B$ en esas posiciones tienen iguales intensidades de grises, lo cual es erróneo.

2. *Valor Absoluto*: Considerar el valor absoluto de todos los elementos de la imagen diferencia. El problema del Ejemplo 1 queda resuelto de la siguiente forma:
$$\begin{pmatrix} 8 & 3 & 2 \\ 9 & 15 & 1 \\ 4 & 7 & 2 \end{pmatrix} - \begin{pmatrix} 8 & 1 & 5 \\ 3 & 12 & 2 \\ 6 & 4 & 1 \end{pmatrix} = \begin{pmatrix} 0 & 2 & 3 \\ 6 & 3 & 1 \\ 2 & 3 & 1 \end{pmatrix}$$

   Aunque más intuitivo que el enfoque basado en truncamiento, considerar el valor absoluto provoca que se modifique la frecuencia de algunas intensidades de grises, lo que puede llevar a una variación importante en el histograma de frecuencia.

En general, los enfoques descritos anteriormente no captan la diferencia real que existe entre las imágenes, pues no consideran una ley de adición interna, o sea, no existe una ley aditiva tal que la suma o sustracción de dos imágenes sea una imagen. Por esta razón, es necesario definir una estructura donde las operaciones entre las imágenes sean internas.

## III. ESTRUCTURA ALGEBRAICA EN LAS IMÁGENES

Basados en que los píxeles de las imágenes toman valores enteros positivos es intuitivo considerar una estructura de grupo en $\mathbb{Z}$, y tomar la ley de este grupo para extender de ser posible dicha estructura al campo de las imágenes. Uno de los ejemplos clásicos de grupo en $\mathbb{Z}$ se muestra en el Ejemplo 2. Para más detalles ver [12, 13, 14].

**Ejemplo 2** La siguiente ecuación establece una relación de equivalencia en $\mathbb{Z}$:
$$a \sim b \Leftrightarrow a \equiv b \ (mód \ n); \ a,b \in \mathbb{Z}$$

donde $a \equiv b \ (mód\ n)$ denota que $a$ es congruente con $b$ en módulo $n$. De lo anterior se obtiene que todo elemento de $\mathbb{Z}$ pertenece a alguna de las siguientes clases de equivalencia, $\{C_0, C_1, \ldots, C_{n-1}\}$, o lo que es lo mismo, $\mathbb{Z} = \bigcup_{i=0}^{n-1} C_i$. Un subgrupo invariante o distinguido de $\mathbb{Z}$ es:

$$n\mathbb{Z} = \left\{z \in \mathbb{Z} : \frac{z}{n} \in \mathbb{Z}\right\}.$$

Notar que $n\mathbb{Z}$ es el grupo de los enteros divisibles por $n$. Teniendo en cuenta las observaciones anteriores es posible demostrar que existe el grupo cociente $\frac{\mathbb{Z}}{n\mathbb{Z}}$ definido por la ley $C_a + C_b = C_{a+b}$, y que además este grupo es abeliano [12, 13, 14]. En lo adelante el grupo cociente $\frac{\mathbb{Z}}{n\mathbb{Z}}$ se denotará como grupos $\mathbb{Z}_n$.

Una de las principales ventajas que tiene el grupo $\mathbb{Z}_n$ es que al operar con cualquier dos elementos de $\mathbb{Z}$ el resultado está en el intervalo $[0; n-1]$, o sea, solo toma valores positivos. Esta característica es favorable si se consideran las intensidades de los niveles de grises de las imágenes como elementos del grupo $\mathbb{Z}_n$, pues no se generan valores negativos al sustraer dos imágenes, y por lo tanto, se evitan los problemas presentados en la Sección II.

El siguiente resultado se obtiene al considerar $G_{k \times m}(\mathbb{Z}_n)$ como el conjunto de imágenes de dimensión $k \times m$, donde el valor de intensidad de cada pixel se toma en el grupo $\mathbb{Z}_n$. La prueba del mismo se puede consultar en [20].

**Teorema 1** El conjunto de imágenes $(G_{k \times m}(\mathbb{Z}_n), +)$, donde la operación aditiva $(+)$ se hereda del grupo $\mathbb{Z}_n$, definida pixel por pixel, tiene estructura de grupo abeliano.

El resultado anterior garantiza que el conjunto de las imágenes de dimensión $k \times m$ tiene estructura de grupo, de forma que existe una ley interna tal que al operar con dos imágenes siempre se obtiene una imagen. Aplicando la ley aditiva del grupo $(G_{k \times m}(\mathbb{Z}_{256}), +)$ para calcular la diferencia entre las imágenes del Ejemplo 1, se obtiene:

$$\begin{pmatrix} 8 & 3 & 2 \\ 9 & 15 & 1 \\ 4 & 7 & 2 \end{pmatrix} - \begin{pmatrix} 8 & 1 & 5 \\ 3 & 12 & 2 \\ 6 & 4 & 1 \end{pmatrix} = \begin{pmatrix} 0 & 2 & 253 \\ 6 & 3 & 255 \\ 254 & 3 & 1 \end{pmatrix}.$$

Con esta nueva estructura de grupos los problemas existentes para la resta de imágenes quedan erradicados, o sea, la imagen resultante de la resta capta las diferencias reales que existen entre las imágenes contrastadas, de forma que no se pierde información o se generan frecuencias indebidas.

La Figura 2 es un ejemplo de la transformación que sufre el histograma de frecuencia de una imagen al restar una imagen escalar $S$ cuyas intensidades de grises en todos los píxeles son 100. En la Figura 2(b) se observa el efecto que produce el truncamiento en la imagen diferencia. Nótese como hay una alta frecuencia en el valor cero mientras que se ha perdido prácticamente toda la información con respecto al histograma de la imagen real (Figura 2(a)). Por otra parte, en la Figura 2(c) se evidencia como el histograma original queda transformado pues se generan nuevas intensidades de grises y aumenta la frecuencia de ocurrencia de algunos píxeles. Sin embargo, cuando se toma en cuenta la ley de grupo propuesta en el Teorema 1, es posible apreciar como el histograma de la imagen original no se transforma, solo se produce un corrimiento uniforme de 100 unidades en los niveles de grises (Figura 2(d)), o sea, se conserva la frecuencia y cantidad de niveles de grises en la imagen resultante.

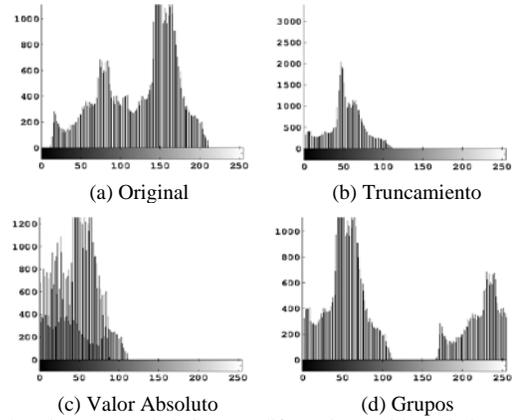

(a) Original  (b) Truncamiento
(c) Valor Absoluto  (d) Grupos

Figura 2. Histogramas de la imagen diferencia considerando la teoría presentada.

El siguiente resultado establece que la cantidad de niveles de grises en una imagen no varía cuando se considera la imagen inversa en el grupo.

**Teorema 2** La entropía de toda imagen $A$ en el grupo $(G_{k \times m}(\mathbb{Z}_n), +)$ es igual a la entropía de su imagen inversa $-A$.

**Demostración:** Una consecuencia de la unicidad del inverso en un grupo es justamente que dos elementos $a$ y $b$ no pueden tener igual inverso. Esta afirmación se demuestra suponiendo que si $a$ y $b$ tienen inverso $c$, entonces el inverso de $c$ es $a$ y $b$, pero por la misma unicidad del inverso se obtiene que $a = b$.

Por otra parte, la entropía de una imagen solo depende de la cantidad de niveles de grises y de la frecuencia de cada nivel de gris, por lo que, como el inverso es único y no existen dos elementos diferentes en el grupo con igual inverso se tiene que la imagen inversa de cualquier imagen $A$ tiene la misma cantidad de niveles de grises y la misma frecuencia, por lo que $E(A) = E(-A)$. ∎

En lo que sigue la notación $A + B$ denota la suma de las imágenes $A$ y $B$ considerando la ley aditiva del grupo $(G_{k \times m}(\mathbb{Z}_n), +)$.

*A. Equivalencia fuerte de imágenes*

En la Figura 2 se mostró la transformación que sufre el histograma de frecuencia al restar una imagen escalar usando las estrategias de truncamiento, valor absoluto y teoría de grupos. Al advertir que solo bajo la operación en el grupo se obtiene el mismo histograma de frecuencia de la imagen real, pero con una traslación uniforme de los niveles de grises, se está en presencia de una propiedad de gran importancia la cual da lugar a la siguiente definición.

**Definición 3.2** (Equivalencia Fuerte) Dos imágenes $A, B \in (G_{k \times m}(\mathbb{Z}_n), +)$ se dicen fuertemente equivalentes $(A \cong B)$ si:

$$A = B + S,$$

donde $S$ es una imagen escalar.

Notar que las imágenes fuertemente equivalentes pueden tener una apariencia visual diferente y sin embargo contienen la misma información. Por lo tanto, para tareas de procesamiento de imágenes pueden ser tratadas como iguales. Mediante esta idea se obtiene el siguiente resultado.

**Teorema 3:** La relación $A \cong B \Leftrightarrow A = B + S$ es de equivalencia en el campo de las imágenes.

**Demostración:**

- *Reflexividad*: El elemento neutro en el grupo es una imagen escalar donde todos los pixeles son justamente el neutro en el grupo $\mathbb{Z}_n$. De aquí que $\forall A \in (G_{k \times m}(\mathbb{Z}_n), +)$ se tiene:
$$A + O = A \Rightarrow A \cong A.$$

- *Simetría*: Sean $A, B \in (G_{k \times m}(\mathbb{Z}_n), +)$ tal que $A = B + S$. La imagen inversa de $S$ es también una imagen escalar y cumple:
$$B = A + (-S) \Rightarrow B \cong A.$$

- *Transitividad*: Sean $A, B, C \in (G_{k \times m}(\mathbb{Z}_n), +)$ donde $A \cong B \wedge B \cong C$. Lo anterior implica que existen imágenes escalares $S_1, S_2$ tales que:
$$A = B + S_1 \wedge B = C + S_2.$$
Sustituyendo la expresión de la imagen $B$ en la ecuación de $A$ se obtiene:
$$A = C + S_1 + S_2,$$
pero $\hat{S} = S_1 + S_2$ es una imagen escalar, de aquí que $A = C + \hat{S} \Rightarrow A \cong C$. ∎

En la Definición 1.1 se establece el concepto de equivalencia débil entre imágenes, y se demuestra que la misma es una caracterización del criterio de parada del algoritmo $MSHi$. El siguiente resultado establece la relación que existe entre los conceptos de equivalencia débil y fuerte.

**Teorema 4:** Si dos imágenes son fuertemente equivalentes entonces son débilmente equivalentes.

**Demostración:** Sean $A$ y $B$ dos imágenes fuertemente equivalentes, o sea, $A = B + S$. Como la imagen $B + S$ tiene los mismos niveles de grises y frecuencias que $B$ se obtiene que $E(A) = E(B + S) = E(B)$, lo que implica que $A$ y $B$ son débilmente equivalentes. ∎

Notar que el recíproco no es cierto, basta considerar dos imágenes con iguales frecuencias de niveles de grises tal que una de ellas no se pueda obtener mediante una traslación uniforme de las intensidades de la otra. Un contraejemplo sencillo se puede apreciar en las imágenes de la Figura 3.

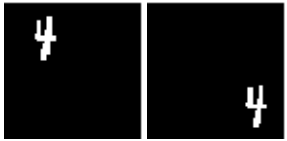

Figura 3. Imágenes con igual entropía que no son fuertemente equivalentes.

*B. Grupo cociente*

El hecho de que las imágenes fuertemente equivalentes tengan la misma información, induce de manera natural la idea de considerar las clases de equivalencia obtenidas mediante la relación de equivalencia fuerte.

Para toda imagen $A$, la clase de equivalencia asociada a dicha imagen se denota como $C_A$ y está dada por:
$$C_A = \{A + S: S \in N\},$$
donde $N$ denota el conjunto de las imágenes escalares, o sea,
$$N = \{S \in (G_{k \times m}(\mathbb{Z}_n), +): S \text{ es una imagen escalar}\}.$$

Mediante las clases de equivalencia se tiene una nueva estructura en el espacio de las imágenes, en el sentido que imágenes con iguales propiedades quedan agrupadas en un mismo conjunto. Uno de los aspectos más importantes al considerar las clases de equivalencia es probar la existencia del grupo cociente en el espacio de las imágenes.

**Teorema 5**: El conjunto cociente $\frac{(G_{k \times m}(\mathbb{Z}_n), +)}{N}$ tiene estructura de grupo.

**Demostración**: La prueba se realiza en dos pasos, primero es necesario demostrar que $N$ es un subgrupo de $(G_{k \times m}(\mathbb{Z}_n), +)$ y luego que es invariante.

1) *$N$ es un subgrupo de $(G_{k \times m}(\mathbb{Z}_n), +)$*:
   a) Para cualquier $S_1, S_2 \in N$ se tiene que $S_1 + S_2$ es una imagen escalar cuyo valor en todos los píxeles es $s_1 + s_2$, donde $s_1$ es el único valor de intensidad de la imagen $S_1$ y $s_2$ es el valor asociado a la imagen $S_2$. De aquí que $S_1 + S_2 \in N$.
   b) Sea $S \in N$. Todos los píxeles de la imagen inversa de $S$ tienen valor $-s$, donde $s$ es la intensidad de los píxeles de $S$, y $-s$ es el inverso de $s$ en el grupo $\mathbb{Z}_n$. Por definición, al tener la imagen inversa de $S$ todos sus píxeles iguales se obtiene que $-S \in N$.

2) *$N$ es invariante*: Para demostrar este paso es necesario apoyarse en el Teorema del Subgrupo Invariante [12]. Mediante este resultado se obtiene que $N$ es invariante si y solo si para todo imagen $A \in (G_{k \times m}(\mathbb{Z}_n), +)$ y $S \in N$, se cumple que $A + S + (-A) \in N$.
   a) Como el grupo $(G_{k \times m}(\mathbb{Z}_n), +)$ es abeliano se tiene que:
   $$A + S + (-A) = \big(A + (-A)\big) + S = O + S = S \in N$$
   donde $O$ denota el neutro del grupo $(G_{k \times m}(\mathbb{Z}_n), +)$.
   Lo anterior demuestra que $N$ es invariante. ∎

Una de las ventajas del uso del grupo cociente en el espacio de las imágenes es que se puede trabajar con cualquier representante de las clases de equivalencia, o sea, dado $C_A \in \frac{(G_{k \times m}(\mathbb{Z}_n), +)}{N}$, cualquier imagen $A_1 \in C_A$ es representativa de la imagen $A$. Este tipo de libertades puede ser valiosa en algunas tareas del procesamiento de imágenes, y para esta investigación en específico, permite definir formalmente un nuevo índice de similitud entre imágenes.

**Definición 3.3** (Distancia Natural de Entropía ($NED$)):
Sean $C_A, C_B \in \frac{(G_{k \times m}(\mathbb{Z}_n), +)}{N}$, $A_1 \in C_A$ y $B_1 \in C_B$ dos imágenes representantes de su clase. La distancia natural de entropía entre $A_1$ y $B_1$ se define como:
$$\hat{v}(A_1, B_1) = E(A_1 - B_1).$$

En la siguiente sección se estudiarán algunas de las principales propiedades de $NED$.

## IV. PROPIEDADES DE LA DISTANCIA NATURAL DE ENTROPÍA

Una de las principales ventajas que tiene el nuevo índice de similitud es que toma valores en un intervalo cerrado y acotado, específicamente
$$\hat{v}(A_1, B_1) \in [0, \log_2 \mathfrak{B}],$$
donde $\mathfrak{B}$ es la cantidad de bits por píxeles en la imagen. La no negatividad y acotación de $NED$ se obtiene de manera inmediata pues la entropía es una función no negativa con máximo en $\log_2 \mathfrak{B}$. Notar que en este punto es muy sencillo normalizar el índice $NED$, basta considerar
$$\frac{\hat{v}(A_1, B_1)}{\log_2 \mathfrak{B}} \in [0, 1],$$

tal que, para valores cercanos a cero, mayor es la similitud entre las imágenes contrastadas, mientras que para valores próximos a uno, se tiene gran discrepancia entre dichas imágenes.

Otro de las características importantes de este nuevo índice de similitud es que cumple la propiedad de simetría, este resultado se propone en el siguiente teorema.

**Teorema 6**: Para toda imagen $A$, $B$ se cumple que:
$$\hat{v}(A,B) = \hat{v}(B,A).$$

**Demostración**: Sean $A \in C_A, B \in C_B$ dos imágenes. Considerando $NED$ se tiene que:
$$\begin{aligned}\hat{v}(A,B) &= E(A-B) = E(-(A-B)), \text{por Teorema 2}\\ &= E(-A+(-(-B))) = E(-A+B)\\ &= E(B+(-A)) = E(B-A), el\ grupo\ es\ abeliano\\ &= \hat{v}(B,A).\blacksquare\end{aligned}$$

El siguiente resultado muestra que $NED$ cumple la identidad de los indiscernibles.

**Teorema 7**: Para toda imagen $A_1 \in C_A \wedge B_1 \in C_B$ se cumple:
$$\hat{v}(A_1,B_1) = 0 \Leftrightarrow C_A = C_B.$$

**Demostración**: La demostración de la necesidad se logra partiendo de que la entropía solo es cero en una imagen cuando todos los píxeles tienen la misma intensidad de nivel de gris, o lo que es lo mismo, cuando la imagen es escalar, de aquí que:
$$\begin{aligned}\hat{v}(A_1,B_1) = 0 &\Rightarrow A_1 - B_1 = S\ (S\ imagen\ escalar)\\ &\Rightarrow A_1 \cong B_1 \Rightarrow C_A = C_B.\end{aligned}$$

Para demostrar la suficiencia se considera $A_1 \in C_A \wedge B_1 \in C_B$, donde $C_A = C_B$. Como $A_1$ y $B_1$ pertenecen a la misma clase se tiene que $A_1 = B_1 + S$, por lo tanto:
$$\begin{aligned}\hat{v}(A_1,B_1) &= E(A_1 - B_1) = E((B_1+S)-B_1), grupo\ abeliano\\ &= E((B_1-B_1)+S) = E(O+S) = E(S) = 0,\end{aligned}$$
lo que demuestra el enunciado. $\blacksquare$

El índice $NED$ se puede programar fácilmente y el costo computacional del mismo es de $\Theta(1)$, por lo que es muy eficiente para obtener una medida rápida de la diferencia existente entre dos imágenes de cualquier tamaño y resolución. Las buenas propiedades tanto analíticas como computacionales que tiene este nuevo índice de similitud, justifican que sea factible incluirlo en los principales algoritmos de análisis de imágenes como un criterio fiable para calcular la similitud entre dos imágenes.

Una de las motivaciones fundamentales de este trabajo era establecer un criterio de parada adecuado para el "Algoritmo Iterativo de la Media Desplazada", o sea, proponer un nuevo índice de similitud entre imágenes que fuese capaz de captar las diferencias que existen entre las imágenes generadas mediante $MSHi$. La vinculación de $NED$ con el algoritmo $MSHi$ es sencilla al considerar la diferencia entre las imágenes de la iteración $k$ y $k-1$, de forma que el algoritmo se detiene cuando
$$\hat{v}(A_k, A_{k-1}) \leq \varepsilon,$$
donde $\varepsilon$ es el umbral de parada fijado de acuerdo al nivel de segmentación que se desee obtener. Establecer el índice $NED$ como nuevo criterio de parada es conveniente pues en cada iteración del algoritmo $MSHi$ se computa la similitud entre dos imágenes, por lo que es necesario un índice que sea rápido y eficiente computacionalmente. En el Algoritmo 1 se propone el esquema general de $MSHi$ usando $NED$ como criterio de parada.

---

**Algoritmo 1**: Algoritmo Iterativo de la Media Desplazada ($MSHi$)

---

**Datos**: Imagen $A$, vector de parámetros $h = (h_r, h_s)$, tolerancia $\varepsilon$.
**Resultado**: Imagen segmentada $B$.
1- **Inicialización**: $errabs = \infty$.
2- **Mientras** $\varepsilon < errabs$ **hacer**
   a. Filtrar la imagen $A$ mediante la Media Desplazada, se obtiene imagen $B$;
   b. Calcular $errabs = E(A-B)$;
   c. Hacer $A = B$;
3- **Retornar** $B$.

---

En lo que sigue, se entiende por $MSHi_{NED}$ al uso de $NED$ como criterio de parada del algoritmo $MSHi$. En el caso de usar el criterio de parada dado en la expresión (1) se emplea la notación $MSHi_{WE}$.

## V. Resultados numéricos

En esta sección se propone una serie de experimentos numéricos que permiten validar la distancia natural de entropía como criterio de parada del algoritmo $MSHi$. Con este objetivo se emplearán imágenes extraídas de la base de datos de Berkeley [16]. Esta base de datos incluye más de 300 imágenes de prueba y para cada imagen se proponen más de cuatro segmentaciones ideales ("*ground truth*"). La selección de las imágenes de prueba se realiza atendiendo a diferentes características, como existencia de altas y bajas frecuencias, oclusión parcial y tamaño de la imagen. En la Figura 4 se muestran las imágenes seleccionadas.

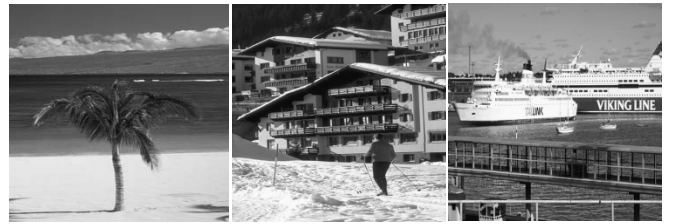

(a) Imagen $\textbf{\textit{No.}}\ \textbf{46076}$    (b) Imagen $\textbf{\textit{No.}}\ \textbf{61086}$    (c) Imagen $\textbf{\textit{No.}}\ \textbf{78019}$

Figura 4. Imágenes de prueba obtenidas de la base de datos de Berkeley.

Todos los experimentos que se presentan en esta sección se basan en las siguientes consideraciones:
1- Los tamaños de ventana o ancho de banda son, en todos los experimentos, $hr = 15, hs = 12$. Esta consideración se basa en que los autores del algoritmo $MSHi$ utilizan estos parámetros para las corridas experimentales que proponen en los trabajos [7, 9, 10, 11]. Está claro que variar los tamaños de ventana puede influir en el resultado final de la segmentación, pero estudiar este comportamiento no es objetivo de esta investigación y por demás fue tratado en [8].
2- El umbral de parada usado en el algoritmo $MSHi_{WE}$ es $\varepsilon = 0.5$, mientras que para $MSHi_{NED}$ se considera $\varepsilon = 0.002$. La elección de dichos umbrales se basa en conocimientos previos adquiridos de forma empírica, y aunque para trabajos futuros se impone la necesidad de considerar una nueva alternativa para seleccionar de forma automática un umbral de parada, dicho aspecto no es objetivo de la presente investigación.

### A. Estabilidad del algoritmo $MSHi_{NED}$

Partiendo del hecho que el "Algoritmo Iterativo de la Media Desplazada" es convergente (ver [6, 10]), se espera que un criterio de parada adecuado describa un curva suave y decreciente al graficar las iteraciones realizas contra el valor obtenido por el criterio de parada en cada iteración.

La Figura 5 muestra los gráficos de iteraciones para los algoritmos $MSHi_{NED}$ y $MSHi_{WE}$ considerando las imágenes No. 46076, 61086 y 78019. Al tomar el índice $NED$ como criterio de parada del algoritmo $MSHi$, se observa como las curvas que describen el proceso de segmentación de las imágenes son suaves y monótona decrecientes (Figuras 5(a), 5(b), 5(c)). Lo anterior demuestra experimentalmente que mediante $NED$ se logra percibir la similitud que existe entre las imágenes generadas durante el proceso de segmentación. Es importante recordar que la segmentación de imágenes es un problema mal planteado, o sea, no existe una solución única y los buenos resultados de cualquier segmentación dependen de las necesidades del observador. Por esta razón, en el caso del algoritmo $MSHi$, un buen criterio de parada debe ofrecer una medida adecuada para estimar la similitud entre dos imágenes, de forma que sea posible decidir el nivel de segmentación deseado de acuerdo al umbral de parada establecido.

Una situación completamente diferente se obtiene al estudiar el gráfico de iteraciones del algoritmo $MSHi_{WE}$. En las Figuras 5(d), 5(e), 5(f) se puede observar que, para todos los casos, existe una marcada variación en los valores obtenidos por el criterio de parada entre dos iteraciones consecutivas. Este hecho evidencia que el índice de similitud definido en la expresión (1) no es adecuado pues provoca inestabilidad en el proceso de segmentación. Esta situación puede generar sobresegmentación en la imagen tratada, lo cual conlleva además a un mayor costo computacional debido al gran número de iteraciones que se realiza, o por el contrario, que con pocas iteraciones el algoritmo se detenga.

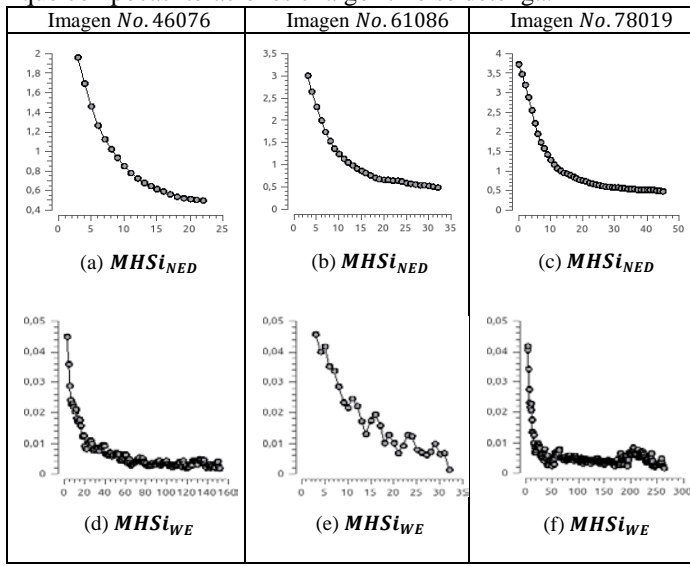

Figura 5. Comportamiento del criterio de parada de los algoritmos $MSHi_{NED}$ y $MSHi_{WE}$.

Una de las inquietudes que puede surgir en esta fase experimental es la diferencia entre los umbrales de parada usados para los algoritmos $MSHi_{NED}$ ($\varepsilon = 0.5$) y $MSHi_{WE}$ ($\varepsilon = 0.002$). En este sentido, la característica más importante a tener en cuenta es que tanto $NED$, como el índice de similitud usado originalmente en el algoritmo $MSHi$, toman valores en el intervalo $[0, 8]$, lo cual es una consecuencia del uso de la función de entropía de Shannon, y de que en este trabajo las imágenes solo toman $2^8$ intensidades de grises. En los gráficos de iteraciones del algoritmo $MSHi_{WE}$, se puede observar como el valor máximo alcanzado por el criterio de parada es 0.05, lo cual constituye otro indicador para mostrar la ineficiencia de este criterio de parada, pues al obtener un índice de similitud muy pequeño se asume que existe poca diferencia entre las imágenes comparadas. En el caso del algoritmo $MSHi_{NED}$ el mayor valor obtenido está cercano a 4, lo cual es un resultado aceptable debido a que no se ha segmentado prácticamente la imagen. De haberse seleccionado $\varepsilon = 0.5$ para el algoritmo $MSHi_{WE}$, no se hubiese podido estudiar el comportamiento de la curva que describe el criterio de parada durante el proceso de segmentación, pues en todos los casos, solo se tendría una iteración del algoritmo.

En la Tabla 1 se relacionan el número de iteraciones realizadas por cada algoritmo. Es interesante apreciar como para todas las imágenes, excepto el caso de la imagen $No.$ 61086, existe una diferencia notable entre el número de iteraciones realizadas por el algoritmo $MSHi_{NED}$ y $MSHi_{WE}$. Esta situación, provocada por la inestabilidad del criterio de parada que aparece en la expresión (1), aumenta de forma considerable el costo computacional y el tiempo de ejecución. Notar como para la imagen $No.$ 78019 existe una diferencia de 218 iteraciones. Un caso de prueba interesante es la imagen $No.$ 61086, en la cual para los umbrales establecidos se obtiene igual cantidad de iteraciones en ambos algoritmos.

|  | Imágenes | | |
|---|---|---|---|
| Algoritmo | 46076 | 61086 | 78019 |
| $MSHi_{NED}$ | 44 | 33 | 46 |
| $MSHi_{WE}$ | 153 | 33 | 264 |

Tabla 1. Número de iteraciones realizadas por los algoritmos $MSHi_{NED}$ y $MSHi_{WE}$.

### B. Segmentación mediante $MSHi_{NED}$

El objetivo fundamental de esta sección es estudiar las segmentaciones que se obtienen mediante el algoritmo $MSHi_{NED}$. Con este fin se emplean las segmentaciones ideales "*ground truth*" (GT) y los índices de similitud "Rank Index" (RI) [17], "Probabilistic Rank Index" (PRi) [18] y "Normalize Probabilistic Rank Index" (NPRi) [19]. Respecto a los índices de similitud es importante tener en cuenta que:

- **RI**: Hace uso de una imagen "*ground truth*", toma valores en el intervalo $[0; 1]$, y para resultados mayores que $0.5$ se asume que la segmentación es buena. En todos los experimentos presentados se ha tomado el primer "*ground truth*" de la base de datos para hallar este índice.

- **PRi**: Hace uso de todas las imágenes "*ground truth*", toma valores en $[0, 1]$, y al igual que el RI cuando el índice es mayor que $0.5$ se considera que la segmentación es adecuada.
- **NPRi**: Hace uso de todas las imágenes "*ground truth*" de la base de datos para calcular el valor esperado, el índice está comprendido en el rango $[-1; 1]$, y para valores mayores que $0$ se asume que la segmentación es correcta.

La imagen $No. 46076$ presenta detalles importantes y bien marcados, como por ejemplo las nubes, la palma, y las sombras (ver Figura 6(a)), los cuales deben ser identificados por cualquier algoritmo de segmentación. Es interesante notar que visualmente las imágenes de las Figuras 6(b), 6(c) son muy parecidas, a pesar de existir una diferencia considerable entre el número de iteraciones realizadas por el algoritmo $MSHi_{WE}$ y $MSHi_{NED}$ (ver Tabla 1). En el caso del algoritmo $MSHi_{NED}$, se distingue detalles importantes, como por ejemplo la montaña al fondo del agua, las sombras de la palma, e incluso se nota la diferencia que existe en la imagen original respecto a las tonalidades que toma la zona del agua que está cercana a la arena.

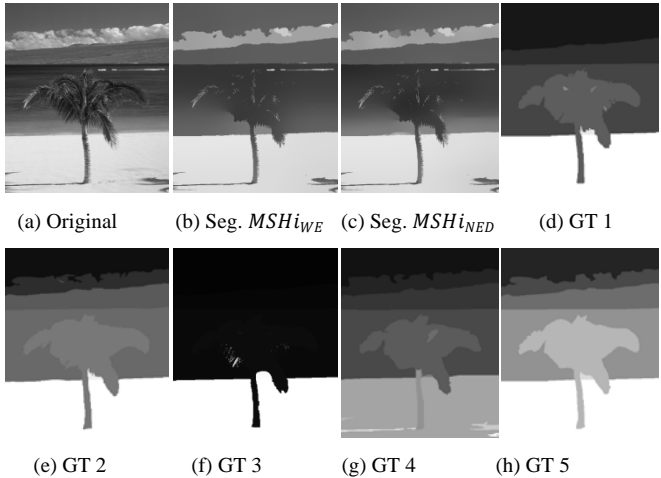

(a) Original   (b) Seg. $MSHi_{WE}$   (c) Seg. $MSHi_{NED}$   (d) GT 1

(e) GT 2   (f) GT 3   (g) GT 4   (h) GT 5

Figura 6. Segmentación de la imagen $No. 46076$.

En la Tabla 1 se hizo notar que, para el caso de la imagen $No. 61086$ los algoritmos $MSHi_{NED}$ y $MSHi_{WE}$ realizan igual número de iteraciones. Está claro que dado este resultado las segmentaciones presentadas en la Figura 7(b) y 7(c) son idénticas, lo cual se obtiene debido a que el algoritmo $MSHi$ es determinista [8, 9, 10]. En general, esta imagen es visualmente complicada pues presenta mucha variabilidad en las intensidades de los niveles de grises, al igual que detalles importantes como las ventanas de los edificios. Es importante notar que mediante la segmentación obtenida por el algoritmo $MSHi_{NED}$ se conservan detalles como las ventanas que no son incluidas en las segmentaciones "*ground truth*", al mismo tiempo se identifican objetos como los palos de esquiar, que por sus características (finos y con intensidad muy similar a la de la nieve), son muy complejos de conservar durante el proceso de segmentación.

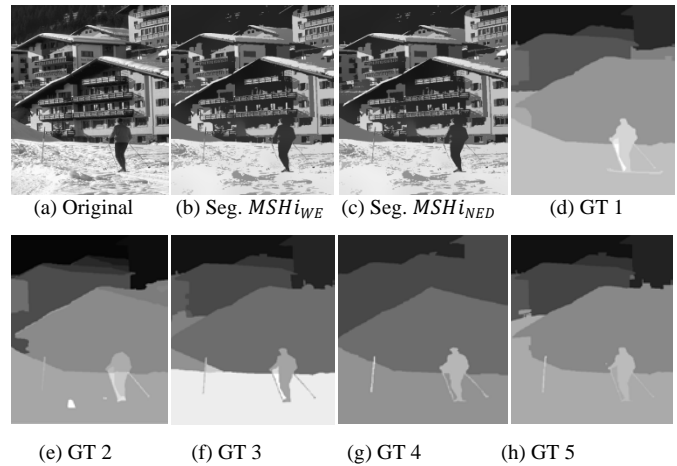

(a) Original   (b) Seg. $MSHi_{WE}$   (c) Seg. $MSHi_{NED}$   (d) GT 1

(e) GT 2   (f) GT 3   (g) GT 4   (h) GT 5

Figura 7. Segmentación de la imagen $No. 61086$.

La imagen de prueba $No. 78019$ (ver Figura 8(a)) es, sin duda, una de las escenas más complicadas desde el punto del análisis visual. La dificultad de la misma radica en la gran cantidad de altas frecuencias espaciales (detalles), algunos de ellos difíciles de discriminar para un observador experimentado en el análisis de esas escenas. Como consecuencia del alto nivel de complejidad de esta imagen y de la inestabilidad del algoritmo $MSHi_{WE}$, se aprecia en la Figura 8(b) como existe sobresegmentación en diferentes lugares de la imagen resultante. Para la segmentación obtenida por el algoritmo $MSHi_{NED}$, puede distinguirse como el cielo queda bien segmentado, al igual que se distinguen perfectamente las letras y ventanas de ambos buques, el aspecto negativo se encuentra en algunas zonas del agua que quedan sin segmentarse de forma adecuada y en el puente, donde se puede apreciar una pequeña fusión de regiones, lo que constituye una situación no deseada.

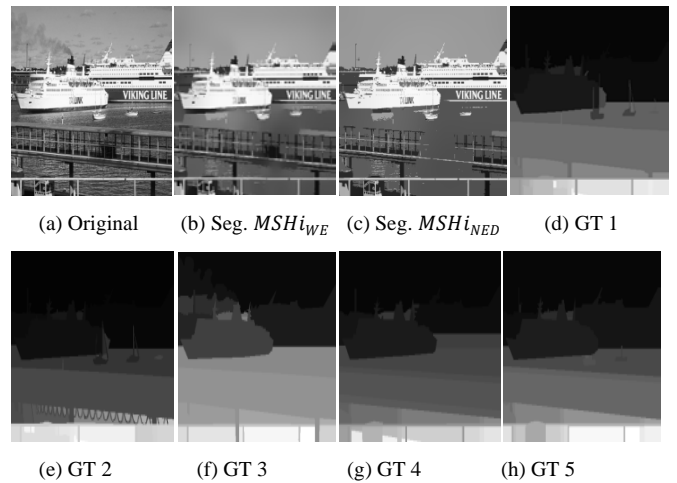

(a) Original   (b) Seg. $MSHi_{WE}$   (c) Seg. $MSHi_{NED}$   (d) GT 1

(e) GT 2   (f) GT 3   (g) GT 4   (h) GT 5

Figura 8. Segmentación de la imagen $No. 78019$.

A pesar que la evaluación visual en la actualidad ha alcanzado un nivel importante, por ser el interpretador quien decide entre una supuesta buena segmentación y otra, no cabe duda que también subyace un análisis subjetivo. Por esta razón, llevar a cabo una evaluación cuantitativa también es necesario. La buena combinación de ambas darán los mejores resultados. Para lograr tal objetivo se emplearon los índices

$RI$, $PRi$ y $NPRi$. En la Tabla 2 se presentan los resultados obtenidos por estos índices de acuerdo a las segmentaciones dadas en las Figuras 6, 7 y 8. La primera observación importante es que para los dos algoritmos estudiados se obtienen altos valores en todos los índices de similitud, lo que evidencia que las segmentaciones obtenidas son adecuadas. A pesar de que no existir gran diferencia entre los resultados de los índices, se aprecia que en el caso del algoritmo $MSHi_{NED}$, los valores son ligeramente superiores con respecto a $MSHi_{WE}$, lo que evidencia otra de las ventajas del nuevo criterio de parada. Es importante puntualizar, con el objetivo de llevar a cabo un análisis más real, que aunque no existe una marcada diferencia entre los valores de los índices, el algoritmo $MSHi_{WE}$ realiza muchas más iteraciones que el nuevo enfoque, lo que implica mayor costo computacional y por tanto, una limitación en el campo de las aplicaciones.

|  | ALGORITMO | |
|---|---|---|
| **Imagen $No. 46076$** | $MSHi_{NED}$ | $MSHi_{WE}$ |
| RI | 0.78524 | 0.78336 |
| PRi | 0.80122 | 0.80073 |
| NPRi | 0.50310 | 0.50180 |
| **Imagen $No. 61086$** | $MSHi_{NED}$ | $MSHi_{WE}$ |
| RI | 0.78207 | 0.78207 |
| PRi | 0.78176 | 0.78176 |
| NPRi | 0.45440 | 0.45440 |
| **Imagen $No. 78019$** | $MSHi_{NED}$ | $MSHi_{WE}$ |
| RI | 0.85137 | 0.84306 |
| PRi | 0.85644 | 0.85241 |
| NPRi | 0.64110 | 0.63100 |

Tabla 2. Evaluación de la segmentación obtenida para los algoritmos $MSHi_{NED}$ y $MSHi_{WE}$.

## VI. CONCLUSIONES

Se determinó un nuevo índice de similitud a partir de una estructura algebraica definida en el campo de las imágenes. En particular, se establece una nueva relación de equivalencia entre imágenes, el cual permite probar la existencia del grupo cociente, definiéndose sobre el mismo la distancia natural de entropía $NED$.

Se demostró las propiedades teóricas de NED, de las cuales algunas de ellas resultaron importantes; como son la no negatividad, acotación, simetría, identidad de los indiscernibles y una relativa insensibilidad bajo pequeñas perturbaciones.
Se probó que las segmentaciones obtenidas mediante el algoritmo $MSHi_{NED}$ son adecuadas en comparación con las segmentaciones ideales presentes en la base de datos de Berkeley. Esto quedó demostrado cuantitativamente al observar los altos índices alcanzados mediante los criterios $RI$, $PRi$ y $NPRi$.

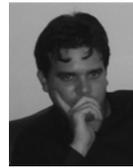
**Yasel Garcés Suárez,** received his bachelor degree in Mathematics from the Havana University in 2011 and the Master degree in 2014. Since 2011, he is part of the Digital Signal Processing Group of the Institute of Cybernetics, Mathematics and Physics (ICIMAF). His research interests include segmentation, restoration, visual pattern recognition, and analysis of images. He has published more than 10 articles in international journals and has participated in many international conferences. He has received more of five national prizes.

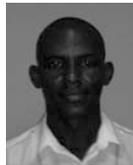
**Esley Torres García**, received his bachelor degree in Mathematics from the Havana University in 2009. Since 2011, he is part of the Digital Signal Processing Group of the Institute of Cybernetics, Mathematics and Physics (ICIMAF). His research interests include segmentation, restoration, visual pattern recognition, and analysis of images. Since 2009, he teaches mathematics at Superior Polytechnic University José Antonio Echavarría (ISPJAE). He has published more than 12 articles in international journals and has participated in many international conferences. He has received one national prize.

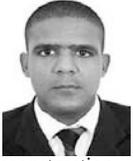
**Osvaldo Pereira Barzaga**. He graduated in Computer Science Engineering in October 2008. He received his degree of Master of Science in mention of Applied Computer in October 2010. Now is part of the Digital Signal Processing Group of the Institute of Cybernetics, Mathematics and Physics (ICIMAF). His research interests include: processing and segmentation of digital images, reconstruction of three-dimensional models from medical imaging, visualization and virtual reality. He is developing his PhD in topics of edge detection in images.

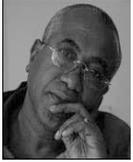
**Roberto Rodríguez M.** received his diploma in Physic from the Physics Faculty, Havana University in 1978 and the PhD degree from Technical University of Havana, in 1995. Since 1998, he is the head of the Digital Signal Processing Group of the Institute of Cybernetics, Mathematics & Physics (ICIMAF). His research interests include Segmentation, Restoration, Mathematical Morphology, Visual pattern recognition, Analysis and Interpretation of images, Theoretical studies of Gaussian Scale-Space and Mean shift. He has published more than 100 articles in international journals and in many international conferences. He has two published books and three written chapters in other related books with the speciality. He has received more of ten prizes national e international.